\documentclass[times, 10pt, twocolumn]{article} 
\usepackage{latex8}
\usepackage{times}
\usepackage{amsmath}
\usepackage{soul,color,graphicx}
\usepackage{multirow}
\usepackage[table]{xcolor}  
\usepackage[hyphens]{url}

\begin{document}

\title{ContamiNet: Detecting Contamination in Municipal Solid Waste}

\author{Khoury Ibrahim\\
Cruise Automation \\ 1201 Bryant Street, San Francisco, CA 94103\\ khoury.ibrahim@getcruise.com\\
\and
Danielle A. Savage\\
First Republic Bank\\
111 Pine Street, San Francisco, CA 94111\\dsavage@firstrepublic.com\\
\and
Addie Schnirel\\
Recology\\
50 California Street, San Francisco, CA 94111\\ 
aschnirel@recology.com\\
\and
Paul Intrevado\\
University of San Francisco\\
2130 Fulton Street, San Francisco, CA 94117\\ 
pintrevado@usfca.edu\\
\and
Yannet Interian\\
University of San Francisco\\
101 Howard Street, San Francisco, CA 94105\\ 
yinterian@usfca.edu\\
}

\maketitle
\thispagestyle{empty}

\begin{abstract}
Leveraging over 30,000 images each with up to 89 labels collected by Recology---an integrated resource recovery company with both residential and commercial trash, recycling and composting services---the authors develop ContamiNet, a convolutional neural network, to identify contaminating material in residential recycling and compost bins. When training the model on a subset of labels that meet a minimum frequency threshold, ContamiNet preforms almost as well human experts in detecting contamination (0.86 versus 0.88 AUC). Recology is actively piloting ContamiNet in their daily municipal solid waste (MSW) collection to identify contaminants in recycling and compost bins to subsequently inform and educate customers about best sorting practices.
\end{abstract}

\Section{Introduction}
\label{intro}
When countries in the Global North make significant policy decisions, the repercussions of those actions can reverberate across the globe, often having a disproportionately negative impact on countries from the Global South. Waste management policy can have a particularly grievous impact. The United States (US) is the largest global producer of municipal solid waste (MSW), which comprises landfill, recycling and compost. The US generates 259 million tons of MSW annually \cite{worldBank2018_01}, more than 50\% of which is recyclable material \cite{umich2018_01}. American domestic waste policy therefore has an outsize global impact.

Heretofore, China has been a common destination for the world's recyclable material. Over several decades, China has developed an industry focused on the ingestion and processing of global recyclable materials. In 2016, the US exported almost 700,000 tons of recycling to China \cite{npr2019_01}. The US, however, isn't the only global culprit: roughly 70\% of global plastic waste---seven million tons---is processed by China annually \cite{npr2019_01}. 

It can be economically viable and even highly profitable to buy, sell, and/or process high-quality, non-contaminated recyclable materials. However, issues arise when those materials are contaminated, which then requires humans or machines to sift through the recycling to remove contaminants, a costly and time-intensive endeavor. E.g., a cardboard pizza box or egg container can be recycled if clean (unsoiled), but are considered contaminated and therefore non-recyclable if they are soiled. Moreover, placing soiled items in a recycling container may soil other items, magnifying contamination. 

The complex global supply chain responsible for transferring the majority of the world's recycling to Chinese processing plants ground to a halt in 2018, when China implemented their \emph{National Sword} policy, aimed at reducing the amount of acceptable contamination in recyclable materials by an order of magnitude, from 5\% to 0.5\%. This policy change was an attempt by China to receive and process only high-quality, uncontaminated recycling. But for many countries, including the US, this highly stringent level is difficult achieve \cite{pri2018_01}. 

To the average residential American, this change in waste management policy is likely transparent. This seismic shift in Chinese policy has however, reverberated globally, shifting the economics of how waste management companies operate, with environmentally detrimental consequences. Locally, American waste management companies continue to collect recycling, but, owing to China's stringent contamination rates and a lack of economically viable alternatives, often landfill or incinerate recyclable materials post collection \cite{nyt2018_01}. Globally, certain countries such as India have tried to absorb the overflow of recycling, but they too have also begun to implement additional restrictions \cite{nyt2018_02}. In the first half of 2018, nearly 50\% of plastic waste exported from the US for recycling was shipped to Thailand, Malaysia and Vietnam, all developing countries in the Global South where there exists little to no regulatory framework to process recycling in an environmentally friendly manner \cite{guardian2018_01}.

The global recycling supply chain is currently in flux. Whereas countries such as China have, for decades, accepted and processed recycling materials with substandard contamination rates, standards are rising across all nations, and the current \emph{modus operandi} will invariably give way to a new, more sustainable equilibrium. With US municipal solid waste increasing by almost 75\% since 1980 \cite{umich2018_01}, there is a real need to process recycling and compost in a sustainable fashion, and in a manner that does not have a disproportionately negative impact on countries of the Global South. 

The authors developed a framework that operates at the residential customer level, leveraging a convolutional neural network (CNN) to identify contamination in individual recycling and compost bins from photos taken upon collection. Although still in the pilot phase, when contamination is detected by the CNN, residents will ultimately be notified of the contamination, and be subsequently provided with guidelines and educational materials about recycling and composting. Engaging and educating individual customers is hypothesized to have a lasting and significant change in behavior. The potential to effect change at the source is an exciting opportunity, and one that aligns with several of the United Nations's Envision 2030 (SDGs) goals of sustainable cities and communities, responsible consumption and production, and climate action \cite{un_01}.

In close partnership with Recology---a San Francisco employee-owned integrated resource recovery company that provides communities along the west coast of the United States with both residential and commercial trash, recycling and composting services---the CNN was trained on 30,000 of their multi-labeled images, and was able to detect contamination in recycling and compost at a level similar to expertly-trained humans.

The paper is arranged as follows: $\S$\ref{relWork} discusses related work, $\S$\ref{ContamiNet} outlines the community partnership between the authors and Recology, details the facets of convolutional neural network employed for image recognition, as well as the data employed to train the CNN. $\S$\ref{results} presents the results, and the manuscript concludes with $\S$\ref{conclusions}, a discussion of conclusions and avenues for future research. 

\Section{Related Work}
\label{relWork}

As this research crosses several traditional academic boundaries, the literature is expansive, from deep learning models and computation to image recognition and waste management. In the interest of brevity, the authors have narrowed the focus of related research at the intersection of the image detection of municipal solid waste. 

Although only tangentially related, Rovetta et al.\ \cite{rovetta2009early} embed several sensors in garbage bins in Pudong, China, to help estimate how full a garbage bin is. Bin levels were calculated based on a combination of image processing and digital distance sensors. Ultimately, this data was used to optimize the scheduling of garbage collection trucks. Hannan et al.\ \cite{hannan2012automated} and Arebey et al.\ \cite{arebey2012solid} use gray level aura matrices and gray level co-occurrence matrix feature extraction, respectively, to determine how full a garbage bin is. Shafiqul Islam et al.\ \cite{KB2014} solves a similar problem using a multi-layer perceptron to estimate the amount of waste inside of a bin based on top-down images of the bin. Although certain applications of the aforementioned research attempt to classify material type in an effort to determine the density and therefore, the weight of the garbage, none of these methods were designed to identify specific objects.

Yang \& Thung \cite{yang2016classification} employ both a convolutional neural network (CNN) and support vector machines (SVMs) to evaluate 2,400 images of waste that include garbage/landfill and recyclable materials). Individual items across six classes---glass, paper cardboard, plastic, metal, and trash---were photographed on a white background, with roughly 400--500 photos per class. Several image augmentation techniques were employed including random rotations, brightness variation, translation, scaling, and cropping. The SVM ultimately out performs the CNN in this research, owing in part to issues of hyperparameter tuning, ascribed to the limited size of the training data. Awe, Mengistu, \& Sreedhar \cite{awe2017final} extend Yang \& Thung's work by stitching two to six individual photos used in in Yang \& Thung's research to create 10,000 \emph{virtual} piles of garbage, while reducing the number of classes from six to three: landfill, recycling, and paper. Fast R-CNN was used for classification, resulting in a mean average precision of 0.683. Sakr et. al.\ \cite{Sakr2016} similarly use both a convolutional neural network (CNN) and support vector machines (SVMs) to classify waste as either plastic, paper or metal, using a $256 \times 256$ image. The SVM achieved an accuracy of 94.8\%, whereas the CNN resulted in 83\% accuracy. The algorithms were implemented on a Raspberry pi 3, and were presumably used to classify and physically sort trash via a mechanical system on a conveyor belt. 

Mittal et al.\ \cite{mittal2016spotgarbage} developed an interesting mobile application that allows users to report garbage/debris they deem to be dumped in an inappropriate location to the collection authorities for remediation. The Android application uses client-side software and resources to analyze and determine whether or not a photo taken by a mobile phone contains garbage. The SpotGarbage application employs a CNN called GarbNet to automatically detect and localize garbage in real-world settings. The paper introduces a new annotated dataset, called Garbage In Images (GINI). 450 images of garbage were labeled, and bounding boxes drawn around the garbage in the image. The model is then trained using patches or sub-sections of images, and the final binary prediction of whether or not garbage is identified in an image is a result of the aggregation of an image's sub-sections. Moreover, the application also identifies the areas in the image where garbage is identified. The model is able to classify images with an accuracy of 87.69\%.

Most recently, Prasanna et al.\ \cite{Prasanna2018} provide a survey of current image detection methods used to classify waste for efficient disposal and recycling. The authors create a taxonomy to classify image detection techniques based on shape, reflectance, materials, or algorithm.

This manuscript differentiates itself from the aforementioned research across multiple axes. We employ a large set of 30,000 images of municipal solid waste created and collected by Recology, the designated collection agency for all residential MSW collection in San Francisco. Photos are taken directly at the point of collection (top-down photos of recycling or compost bins), as well as at the main collection site. Each image was painstakingly tagged by Recology's staff of trained experts to identify up 89 different contaminants, with 400 images tagged by four experts. We train and test several CNNs to identify contaminants in residential recycling and compost bins. With contaminants identified, individual customers can then be mailed educational materials in an effort to eliminate misclassifying items, wish-cycling (customers placing things in their recycling bins they wish could be recycled but in fact, are not recyclable), and improve customers' ability to choose the best outlet for their MSW---clean paper can be both recycled and composted, but knowing which option is the ideal is not always clear.

\Section{ContamiNet}
\label{ContamiNet}

\SubSection{Recology: A Community Partnership}
\label{commPart}

Recology's mission is to create a world without waste by developing and discovering sustainable resource recovery practices that can be implemented globally, a vision that dovetails with the spirit of this research. The academic authors were delighted that Recology reached out with this initiative. Partnering with Recology provided the framework and impetus for this research, as well as helping define research parameters, sharpen algorithmic and community-centric objectives, and  leverage access to vast amounts of data required to train the deep learning model in question.

In San Francisco, Recology contracts to collect municipal solid waste (MSW) for all residents and most commercial enterprises. Although it would be significantly cheaper for Recology to landfill all the collected MSW, they process several streams of MSW in an effort operate in an environmentally sustainable fashion. Municipal solid waste is currently collected in three separate collection bins: trash, recycling, and compost (organic waste). 

It is not uncommon for Recology customers---owing to a lack of knowledge or judiciousness---to place MSW in the incorrect collection bin: placing a recyclable material in a trash bin, trash in a recycling bin, or compostable organic material in a trash bin. Customers may also fail to pre-process certain items before placing them in the correct bins. E.g., when food scraps are not removed from their plastic container, what was once a recyclable plastic container is now classified as trash, and sent to the landfill. 

Placing MSW in an incorrect collection bin is a source of great time, effort and cost for Recology, to say nothing of the averse environmental impact. Although the contents of trash bins are not inspected upon collection, the contents of all recycling and organic compost bins are manually inspected and triaged for contamination at their respective sorting facilities. 

\SubSection{Problem Formulation}

Given a set of photos and a pre-defined list of contaminants, e.g., glass bottles, diapers, e-waste, coffee grounds, etc., can a model be trained to detect contamination in a residential recycling or compost bin? The detection of contaminating items in recycling and compost bins can be formulated as a multi-label classification problem. Inputs to the model are images of the contents of recycling and compost bins, $X$, and the output is a vector $y=(y_1, \ldots, y_K)$, where $y_i$ indicates the absence or presence item $i$ ($y_i \in \{0,1\}$), and $K$ represents the total number of unique items  classified. To train the model, we optimize the sum of $K$ binary cross entropy losses, one for each item $i \in \{1, \dots K\}$.

\SubSection{Data}
\label{data}

Recology, investing a significant amount of time and resources in this research, collected a total of 30,780 photos of the contents of residential recycling and compost bins. The photographs were colour, aerial photos, however there were several variations. Some of the photos were taken directly at a customer's home: top-down aerial photos of the contents of a recycling or compost bin with the lid flipped open, as it would look curb-side, prior to pickup. These photos were limited insofar as the ability to identify only the items sitting on the top of the contents inside the bins. Other top-down photos taken at the customer's residence were of the entire contents of the recycling and compost bins, once they were dumped into the collection truck---in the area called the hopper---and prior to being compacted with all the other recycling or compost. This allowed for a clear view of the complete contents of each bin associated with each residence. Lastly, photos from various angles were taken at Recology recycling and compost collection centers. The photos \emph{were not} classified to indicate whether the photos were taken of material collected from recycling bins or from compost bins.

\begin{figure}[h]
 \includegraphics[width=\linewidth]{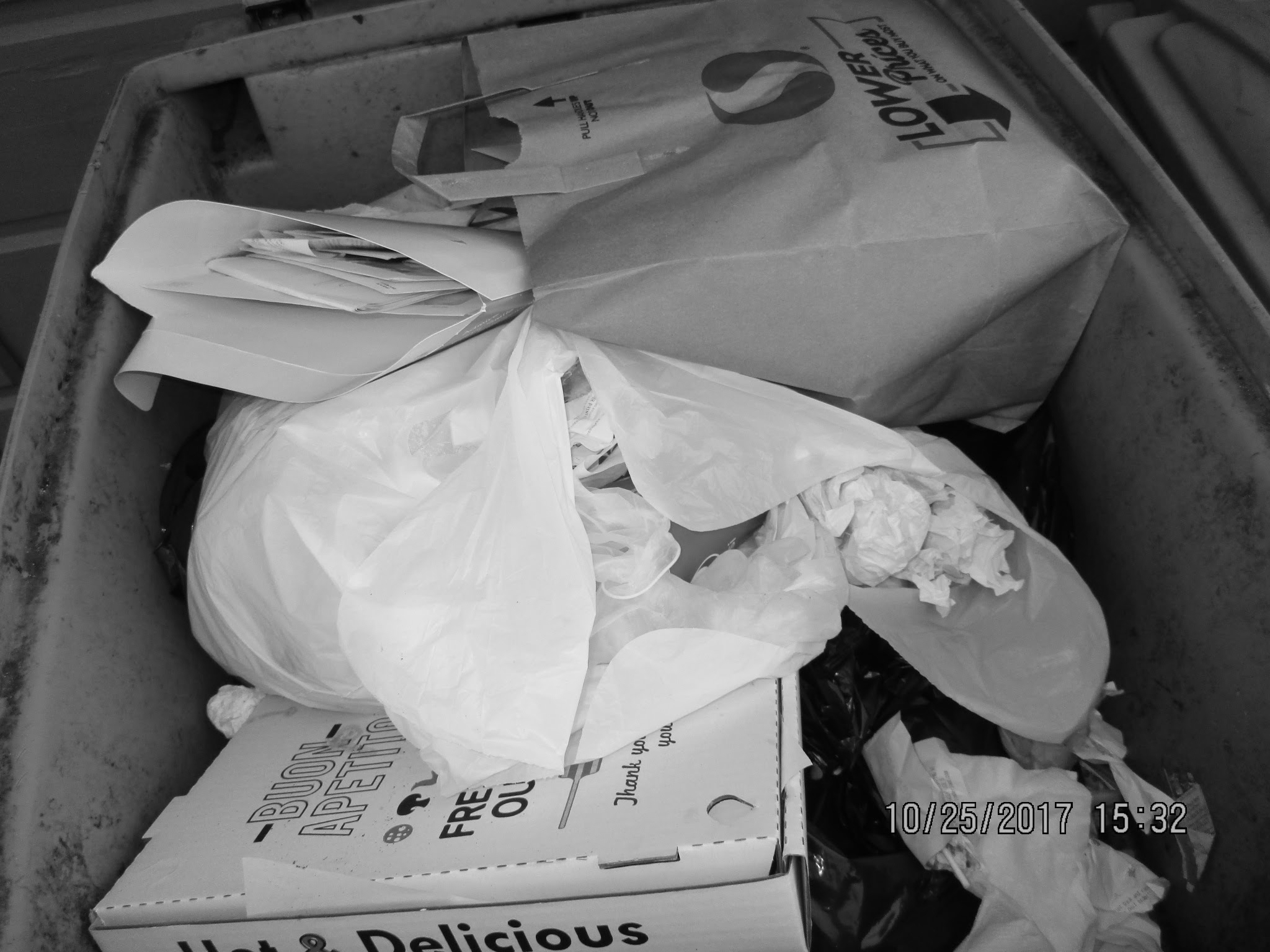} 
\caption{Contents of a Residential Recycling Bin}
\label{fig:recycBin}
\end{figure}

\begin{figure}[h]
\includegraphics[width=\linewidth]{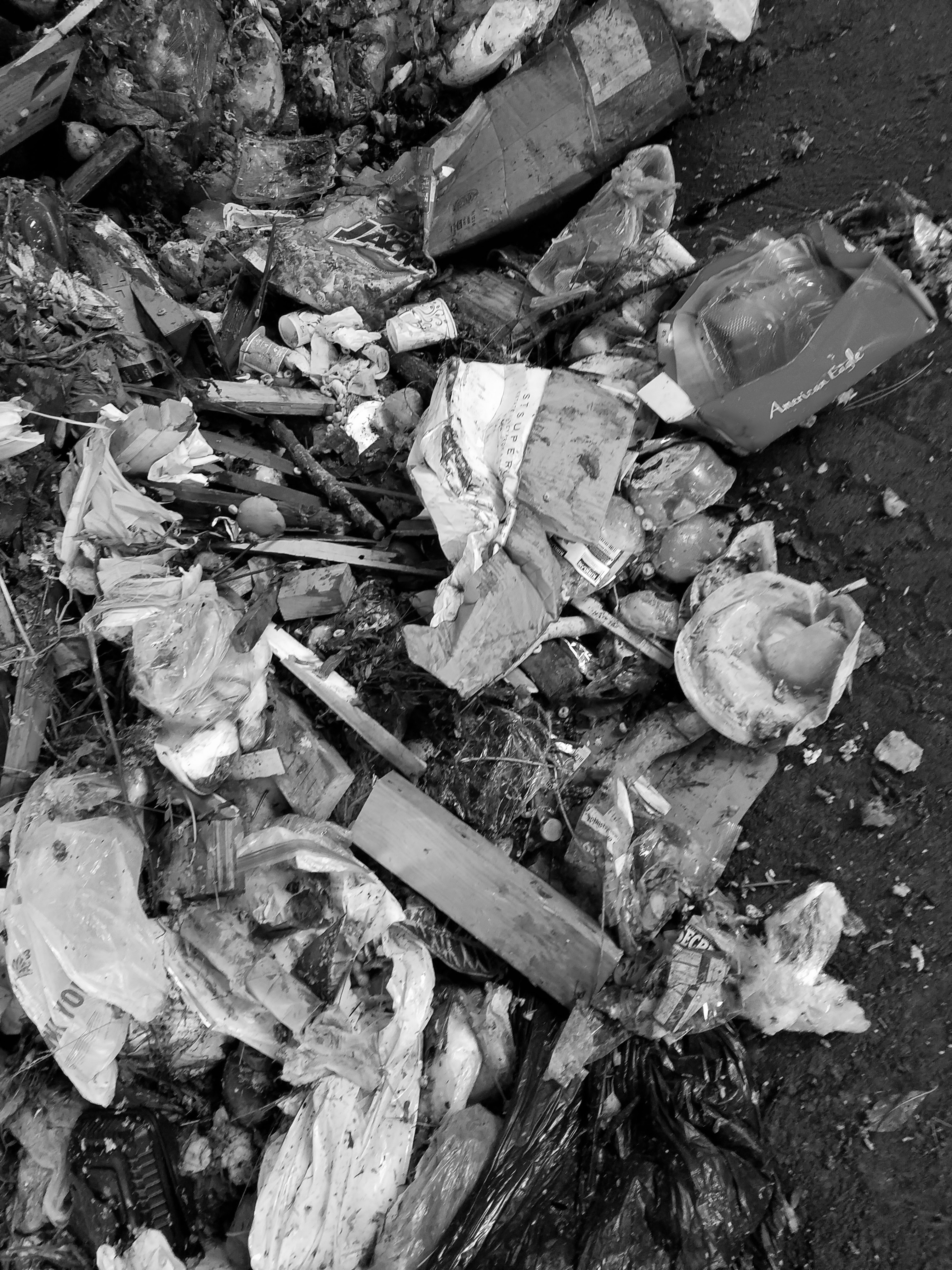} 
\caption{A Heap of Material Collected from Several Compost Bins}
\label{fig:recycHeap}
\end{figure}

Photos of municipal solid waste were taken by Recology employees. There was no established photo protocol regarding angles, photo resolution, or the use of a flash. Most photos were taken with an employee's smart phone or a commodity digital camera. Figures \ref{fig:recycBin} and \ref{fig:recycHeap} are examples of two photos collected by Recology to test the model. Table \ref{table:lblExample} shows the labels each of the four experts associated with the photo in Figure \ref{fig:recycBin}.

\begin{table}[h]
\begin{tabular}{|r|c|c|c|c|}
\hline
\rowcolor[HTML]{EFEFEF} 
 & \multicolumn{4}{c|}{\textbf{Expert}} \\
\hline
\rowcolor[HTML]{EFEFEF} 
\textbf{Labels} & \textbf{1} & \textbf{2}& \textbf{3}& \textbf{4}\\
\hline
 \texttt{paper\char`_flat\char`_clean} & 1	&1	&1	&1 \\
 \rowcolor[HTML]{EFEFEF} 
 \texttt{plastic\char`_film\char`_clean} & 0	&0	&1&	0 \\
 \texttt{paper\char`_flat\char`_soiled} &	0	&0	&1	&0\\
 \rowcolor[HTML]{EFEFEF} 
 \texttt{bagged\char`_items} & 1	&1	&1	&1 \\
 \texttt{paper\char`_bag\char`_clean} & 1	&1&	0&	1 \\
 \rowcolor[HTML]{EFEFEF} 
 \texttt{cart} & 0&	0&	1	&0 \\
 \texttt{paper\char`_napkin\char`_soiled} & 1	&1	&1	&1  \\
 \rowcolor[HTML]{EFEFEF} 
 \texttt{gloves} & 1	&0	&1	&0 \\
 \texttt{cardboard\char`_pizzabox\char`_clean} &1	&1&	1	&1\\
 \rowcolor[HTML]{EFEFEF} 
 \texttt{plastic\char`_bag} &1	&1&	1	&1 \\
 \hline
 \end{tabular}
 \caption{Expert Labels Associated with Figure~\ref{fig:recycBin}}
\label{table:lblExample}
\end{table}

\SubSection{Training the CNN}

The convolutional neural network was trained using 27,342 photos, validated using 3,038 images, and tested using 400 images; test images were labeled by \emph{all} four trained experts to establish inter-rater agreement and a baseline to which model results can be evaluated.

Training images were downscaled to a resolution of $250 \times 333$ pixels, to maintain a constant aspect ratio for all images.  As the model is trained, the images are augmented with random rotations ($\pm10$ degrees), random cropping, and random horizontal flipping. Applying these random image manipulation results in a reduced image size of $234 \times 311$ pixels. Lastly, images are normalized based on the mean and standard deviation of images in the ImageNet \cite{ImageNet} training set. Validation images are  center cropped to  $234 \times 311$ pixels to match the training images, and are similarly normalized. 

Computer vision applications using deep learning are rarely trained from scratch. Leveraging  transfer learning, models are then \emph{fine-tuned} from models that have been trained on ImageNet, for example. In order to use a pre-trained CNN from ImageNet, the last fully-connected layer is removed and replaced with a randomly initialized layer of the appropriate size. Howard and Ruder \cite{HR2018} recently demonstrate a novel idea, whereby combining transfer learning with {\bf discriminative fine-tuning} generates superior results when compared to not implementing discriminative fine-tuning. 

The concept underpinning this approach is to allow each layer to be fine-tuned using different learning rates.  The authors implement a version of discriminative fine-tuning in which layers are divided into three groups:  group one consists of layers closets to the image, group two are middle layers, and group three consists of the newly initialized linear layer. Given a learning rate  $lr$ for group three, learning rates for group 2 and group 1 are set to $lr/3$ and $lr/9$, respectively. This implementation is similar to the one proposed on \url{fast.ai}.

Many deep learning models are trained using a learning rate that decreases over time based on validation loss. The loss function in the validation data set is computed after every epoch: if the loss increases by more than a fixed amount, the learning rate is divided by ten. Training ends either when the maximum number of epochs is reached or the learning rate decreases more than a specified threshold \cite{Ng}. It has been shown \cite{one-cycle} that adaptive learning rate schedules achieve faster convergence and better accuracy.

\begin{figure}[h]
\includegraphics[width=\linewidth]{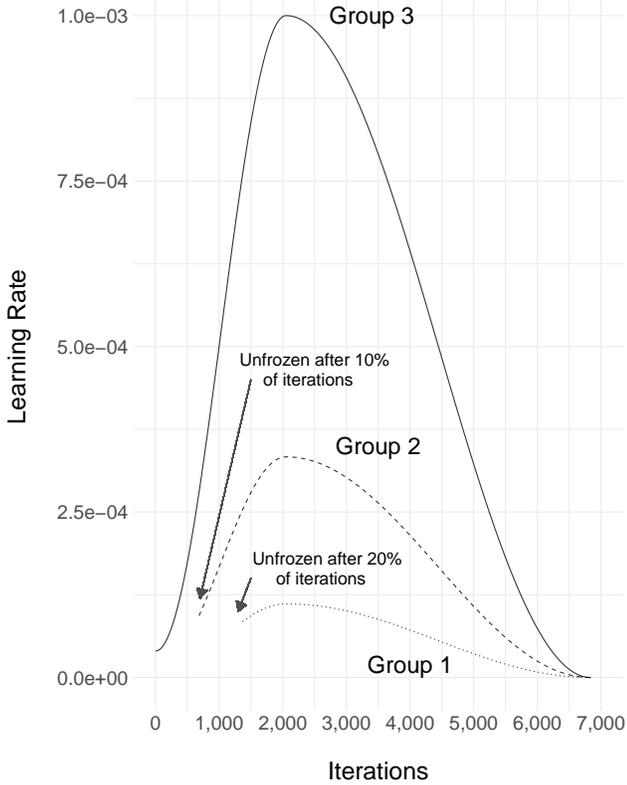}
   \caption{One-Cycle Learning Rates with Cosine Annealing}
   \label{fig:lr}
\end{figure}

This researches implements {\bf one-cycle}  \cite{one-cycle} learning rates with cosine annealing. Given a maximum learning rate $max\_lr$ and the total number of iterations $iters$,  the learning rate is increased using a cosine segment for $30\%$ of the iterations. We subsequently decrease the learning rate according to another cosine segment following the update schedule, seen in Figure \ref{fig:lr}.  The formula for the cosine segment is:

\begin{multline}
lr( lr_1, lr_2, T) = lr_2 + \frac{lr_1 - lr_2}{2} \Bigg(1 + \cos\Big(\frac{i \pi}{T} \Big)\Bigg) \\
\text{for} \hspace{5pt} i \in \{0, \ldots, T-1\}
\label{cosine}
\end{multline}

Using Equation \eqref{cosine}, the  learning rates in Figure \ref{fig:lr} are generated by concatenating the following two cosine segments:

\begin{equation*}
lr\Big(\frac{max\_lr}{25}, max\_lr, (0.3) iters\Big)   
\end{equation*}

and

\begin{equation*}
lr\Big(max\_lr, \frac{max\_lr}{2,000}, (0.3)iters\Big).
\end{equation*}

The network is trained end-to-end using Adam, an algorithm for first-order gradient-based optimization of stochastic objective functions, using standard parameters \cite{KB2014}. The model is trained for 12 epochs using mini-batches of 64 images. The final fully-connected layer is replaced with a layer that contains as many outputs as the number of labels, $K$. Finally, a sigmoid non-linearity is applied to each output. Testing time augmentation is also performed, whereby five versions of each test image are employed using the same transformation as in model training. The predictions are then averaged to generate a final prediction.

\Section{ContamiNet versus the Experts}
\label{results}

To evaluate the quality of ContamiNet, the quality of expert labelers is first assessed. Recall that for the 400 test images, all four experts independently identified contaminants in each of the photos. The area under the receiving operator curve (AUC)---a typical metric used in classification problems---is computed for each expert, by taking the mean of three experts and comparing it to the fourth. This one-versus-the-rest approach is computed for each expert, resulting in four AUC values. A mean AUC is then computed across all experts, by taking the mean of each individual expert's AUC. We similarly computed four AUC's for ContamiNet, one against each of the experts, and the computed the mean AUC for ContamiNet.

In doing so, we observed the mean ContamiNet AUC of of 0.74 to be significantly lower than the mean expert AUC of 0.882. Further exploration uncovered the source of the poor model performance: individual labels in the training data have a highly uneven number of occurrences. Whereas certain labels are identified in almost a third of the 30,780 training images (see Table \ref{table:tags_max}), others, such as \texttt{waste\char`_pet} or \texttt{toxic\char`_batteries} occur only four or six times respectively in the entire set of training images. Table \ref{table:tags_min} shows the labels occurring with the lowest frequency in the training images.

\begin{table}[ht!]
\begin{centering}
\begin{tabular}{|c|c|}
\hline
\rowcolor[HTML]{EFEFEF} 
\textbf{Labels} & \textbf{Frequency}\\
\hline
 \texttt{plastic\char`_bag} &12,049 \\
 \rowcolor[HTML]{EFEFEF} 
 \texttt{bagged\char`_items} & 11,660 \\
 \texttt{cardboard\char`_clean} &7,363\\
 \rowcolor[HTML]{EFEFEF} 
 \texttt{cart} & 6,301 \\
 \texttt{paper\char`_napkin\char`_soiled} & 5,906 \\
 \rowcolor[HTML]{EFEFEF} 
 \texttt{food\char`_scraps} & 5,810 \\
 \texttt{paper\char`_flat\char`_clean} & 5,219 \\
 \rowcolor[HTML]{EFEFEF} 
 \texttt{plastic\char`_film\char`_clean} & 4,932 \\
 \texttt{plastic\char`_rigid\char`_lid} &4,565 \\
 \rowcolor[HTML]{EFEFEF} 
 \texttt{plastic\char`_rigid\char`_bottle} & 3,920 \\
 \hline
 \end{tabular}
 \caption{Ten Most Frequently Identified Labels in Training Images}
\label{table:tags_max}
\end{centering}
\end{table}

\begin{table}[h!]
\begin{centering}
\begin{tabular}{|c|c|}
\hline
\rowcolor[HTML]{EFEFEF} 
\textbf{Labels} & \textbf{Frequency}\\
\hline
 \texttt{toxic\char`_batteries} & 4 \\
 \rowcolor[HTML]{EFEFEF} 
 \texttt{waste\char`_pet} & 6\\
 \texttt{cert\char`_compostable\char`_mini\char`_cup} & 7 \\
 \rowcolor[HTML]{EFEFEF} 
 \texttt{waste\char`_lightbulb} & 8 \\
 \texttt{ewaste\char`_ink\char`_cartridge} & 8 \\
 \hline
 \end{tabular}
 \caption{Five Least Frequently Identified Labels in Training Images}
\label{table:tags_min}
\end{centering}
\end{table}

Deep learning models such as convolutional neural networks need a minimal amount of images to be associated with an individual label, so that it can learn to identify the label with a reasonable degree of accuracy. Owing to the low frequency of occurrence of certain labels, the authors attempted several experiments, setting varying thresholds for label frequency, such that any label occurring less frequently than some threshold be removed the the training images, and therefore, the model.

Choosing minimal label frequency threshold values of 100, 300, and 1,000, the authors trained the CNN using images and labels modified based on the varying threshold values, resulting in significantly improved AUC results (see Table \ref{table:thresholds}). 

\begin{table}[h!]
\begin{centering}
\begin{tabular}{ |c |c | c| c|}
\hline
\rowcolor[HTML]{EFEFEF} 
\textbf{Minimum} & \textbf{\# Unique} & \textbf{Mean} & \textbf{Mean} \\
\rowcolor[HTML]{EFEFEF} 
\textbf{Label } & \textbf{Labels} & \textbf{Expert} & \textbf{ContamiNet} \\
\rowcolor[HTML]{EFEFEF} 
\textbf{Threshold} & & \textbf{AUC} & \textbf{AUC}\\
\hline
100 & 67 & 0.76& 0.87\\
\rowcolor[HTML]{EFEFEF} 
300 & 50 & 0.80 & 0.88\\
1, 000 & 34 & 0.86 & 0.88\\
 \hline
 \end{tabular}
 \caption{Expert and Model AUC with Varying Label Frequency Thresholds}
\label{table:thresholds}
\end{centering}
\end{table}

At a minimal threshold frequency level of 1,000, i.e., the model is trained only on labels that occur 1,000 times or more in the 27,342 training images, ContamiNet AUC is almost equal to that of human AUC. In establishing this threshold of 1,000 for label frequency that results in 34 labels, 89\% of the observations are retained, ensuring the model is still trained by a large majority of the data.

In fact, constructing 95\% confidence intervals (CIs) on 10,000 bootstrapped samples sampled with replacement from the test data, ContamiNet achieves an AUC statistically significantly better than \emph{Expert 3} , and statistically indistinguishable from \emph{Expert 1} (see Table \ref{table:cis}). Only \emph{Experts 2 \& 4} are statistically significantly better than ContamiNet, albeit by a slim margin. Practically speaking, ContamiNet performs at a level equal to that of an expert human.

\begin{table}[h!]
\begin{centering}
\begin{tabular}{| r | c | c|}
\hline
\rowcolor[HTML]{EFEFEF} 
   & \textbf{AUC} &  \textbf{95\% CI} \\
\hline
\emph{Expert 1} &  0.881  & [0.870, 0.892]\\
\emph{Expert 2} &  0.931  & [0.916, 0.948]\\
\emph{Expert 3} &  0.821  & [0.808, 0.834]\\
\emph{Expert 4} &  0.895  & [0.882, 0.907]\\
 \hline
 \rowcolor[HTML]{EFEFEF} 
\textbf{Expert Mean} & 0.882 & [0.875, 0.890]\\
\hline
\rowcolor[HTML]{EFEFEF} 
\textbf{ContamiNet}  & 0.860& [0.854, 0.871]\\
 \hline
 \end{tabular}
 \caption{Comparison of humans and ContamiNet for the dataset with 34 labels. Test AUC are shown together with confidence intervals.}
\label{table:cis}
\end{centering}
\end{table}

\Section{Conclusions \& Future Research}
\label{conclusions}

The ability to identify, educate, and engage residential customers about their recycling habits is akin to crowd-sourcing an otherwise intractable problem. Changing municipal, state, national and international policy on municipal solid waste management will certainly have a significant impact on how sustainably and equitably waste is processed. Those changes, however, tend to be both glacial and highly political. Not willing to stand idly by, Recology is pushing the frontier of MSW collection in their own municipality. Leveraging ContamiNet to identify, educate and alter customer habits, Recology is poised to fundamentally alter how municipal solid waste is collected and processed. 

In this research, the authors have provided a framework in which contaminants of recycling and compost bins can be detected from a set of uncurated photos without a photography protocol, demonstrating a highly-effective level of efficacy almost as good as that of human experts trained to detect said contamination. ContamiNet's performance can be improved if a photography protocol is implemented, higher resolution photos taken, and if photos are labeled as either coming from a recycling bin or compost bin. Additionally, the use of  bounding boxes  can be used to identify specific items in the photos, allowing ContamiNet to learn far faster and, the authors hypothesize, more accurately.

\bibliographystyle{latex8}
\bibliography{latex8}

\begin{thebibliography}{10}\setlength{\itemsep}{-1ex}\small

\bibitem{Prasanna2018}
M.~Adhithya~Prasanna, S.~Vikash~Kaushal, and P.~Mahalakshmi.
\newblock Survey on identification and classification of waste for efficient
  disposal and recycling.
\newblock pages 520--523, 2018.

\bibitem{nyt2018_01}
L.~Albeck-Ripka.
\newblock Your recycling gets recycled, right? maybe, or maybe not, 2018.
\newblock
  \url{https://www.nytimes.com/2018/05/29/climate/recycling-landfills-plastic-papers.html}.

\bibitem{arebey2012solid}
M.~Arebey, M.~Hannan, R.~A. Begum, and H.~Basri.
\newblock Solid waste bin level detection using gray level co-occurrence matrix
  feature extraction approach.
\newblock {\em Journal of environmental management}, 104:9--18, 2012.

\bibitem{awe2017final}
O.~Awe, R.~Mengistu, and V.~Sreedhar.
\newblock Smart trash net: Waste localization and classification.
\newblock {CS229 Project Report}, Department of Computer Science, Stanford
  University, 2017.

\bibitem{worldBank2018_01}
W.~Bank.
\newblock What a waste global database, 2018.
\newblock
  \url{https://datacatalog.worldbank.org/dataset/what-waste-global-database}.

\bibitem{umich2018_01}
U.~o.~M. Center~for Sustainable~Systems.
\newblock Municipal solid waste fact sheet, 2018.
\newblock
  \url{http://css.umich.edu/sites/default/files/Municipal\_Solid\_Waste\_Factsheet\_CSS04-15\_e2018.pdf}.

\bibitem{nyt2018_02}
M.~Corkery.
\newblock Your recycling gets recycled, right? maybe, or maybe not, 2019.
\newblock
  \url{https://www.nytimes.com/2019/03/16/business/local-recycling-costs.html}.

\bibitem{ImageNet}
J.~Deng, W.~Dong, R.~Socher, L.~jia Li, K.~Li, and L.~Fei-fei.
\newblock Imagenet: A large-scale hierarchical image database.
\newblock In {\em In CVPR}, 2009.

\bibitem{guardian2018_01}
T.~Guardian.
\newblock Huge rise in us plastic waste shipments to poor countries following
  china ban, 2019.
\newblock
  \url{https://www.theguardian.com/global-development/2018/oct/05/huge-rise-us-plastic-waste-shipments-to-poor-countries-china-ban-thailand-malaysia-vietnam}.

\bibitem{hannan2012automated}
M.~Hannan, M.~Arebey, R.~A. Begum, and H.~Basri.
\newblock An automated solid waste bin level detection system using a gray
  level aura matrix.
\newblock {\em Waste management}, 32(12):2229--2238, 2012.

\bibitem{HR2018}
J.~Howard and S.~Ruder.
\newblock Universal language model fine-tuning for text classification.
\newblock In {\em Proceedings of the 56th Annual Meeting of the Association for
  Computational Linguistics (Volume 1: Long Papers)}, pages 328--339.
  Association for Computational Linguistics, 2018.

\bibitem{npr2019_01}
C.~Joyce.
\newblock Where will your plastic trash go now that china doesn't want it?,
  2019.
\newblock
  \url{https://www.npr.org/sections/goatsandsoda/2019/03/13/702501726/where-will-your-plastic-trash-go-now-that-china-doesnt-want-it}.

\bibitem{KB2014}
D.~P. Kingma and J.~Ba.
\newblock Adam: {A} method for stochastic optimization.
\newblock In {\em 3rd International Conference on Learning Representations,
  {ICLR} 2015, San Diego, CA, USA, May 7-9, 2015, Conference Track
  Proceedings}, 2015.

\bibitem{pri2018_01}
J.~Margolis.
\newblock Mountains of us recycling pile up as china restricts imports, 2018.
\newblock
  \url{https://www.pri.org/stories/2018-01-01/mountains-us-recycling-pile-china-restricts-imports}.

\bibitem{mittal2016spotgarbage}
G.~Mittal, K.~B. Yagnik, M.~Garg, and N.~C. Krishnan.
\newblock Spotgarbage: smartphone app to detect garbage using deep learning.
\newblock In {\em Proceedings of the 2016 ACM International Joint Conference on
  Pervasive and Ubiquitous Computing}, pages 940--945. ACM, 2016.

\bibitem{un_01}
T.~U. Nations.
\newblock \#envision2030: 17 goals to transform the world for persons with
  disabilities, 2015.
\newblock
  \url{https://www.un.org/development/desa/disabilities/envision2030.html }.

\bibitem{Ng}
P.~Rajpurkar, J.~Irvin, K.~Zhu, B.~Yang, H.~Mehta, T.~Duan, D.~Ding, A.~Bagul,
  C.~Langlotz, K.~Shpanskaya, et~al.
\newblock Chexnet: Radiologist-level pneumonia detection on chest x-rays with
  deep learning.
\newblock {\em arXiv preprint arXiv:1711.05225}, 2017.

\bibitem{rovetta2009early}
A.~Rovetta, F.~Xiumin, F.~Vicentini, Z.~Minghua, A.~Giusti, and H.~Qichang.
\newblock Early detection and evaluation of waste through sensorized containers
  for a collection monitoring application.
\newblock {\em Waste Management}, 29(12):2939--2949, 2009.

\bibitem{Sakr2016}
G.~Sakr, M.~Mokbel, A.~Darwich, M.~Khneisser, and A.~Hadi.
\newblock Comparing deep learning and support vector machines for autonomous
  waste sorting.
\newblock 11 2016.

\bibitem{one-cycle}
L.~N. Smith.
\newblock A disciplined approach to neural network hyper-parameters: Part
  1--learning rate, batch size, momentum, and weight decay.
\newblock {\em arXiv preprint arXiv:1803.09820}, 2018.

\bibitem{yang2016classification}
M.~Yang and G.~Thung.
\newblock Classification of trash for recyclability status.
\newblock {CS229 Project Report}, Department of Computer Science, Stanford
  University, 2016.

\end{thebibliography}

\end{document}